\pdfoutput=1

\documentclass[11pt]{article}
\usepackage{graphicx}
\usepackage{amsmath}

\usepackage{acl2023}

\usepackage{times}
\usepackage{latexsym}
\usepackage{color}
\usepackage[shortlabels]{enumitem}
\usepackage{xspace}
\usepackage{booktabs}
\usepackage{qtree}

\usepackage[T1]{fontenc}

\usepackage[utf8]{inputenc}
\usepackage{xspace}

\usepackage{microtype}

\usepackage{inconsolata}

\title{Covering Uncommon Ground: \\ Gap-Focused Question Generation for Answer Assessment}

\author{Roni Rabin$^{1}$\hspace{1em} Alexandre Djerbetian$^{1}$\hspace{1em} Roee Engelberg$^{1,2}$\hspace{1em} Lidan Hackmon$^{1}$ \\
        \textbf{Gal Elidan$^{1,3}$\hspace{1em} Reut Tsarfaty$^{1,4}$\hspace{1em} Amir Globerson$^{1,5}$} \\
        $^1$ Google Research~~~
        $^2$ Computer Science Dept., Technion \\
        $^3$ Statistics Dept., Hebrew University of Jerusalem \\
        $^4$ Computer Science Dept., Bar-Ilan University \\
        $^5$ Blavatnik School of Computer Science, Tel Aviv University }
\newcommand{\ts}{T_S}
\newcommand{\tc}{T_C}
\newcommand{\acronym}{{GFQ}\xspace}
\newcommand{\genq}{Q_G}

\newcommand{\commentout}[1]{}

\begin{document}
\maketitle
\begin{abstract}
Human communication often involves information gaps between the interlocutors. For example, in an educational dialogue, a student often provides an answer that is incomplete, and there is a gap between this answer and the perfect one expected by the teacher. Successful dialogue then hinges on the teacher asking about this gap in an effective manner, thus creating a rich and interactive educational experience. We focus on the problem of generating such gap-focused questions ({\acronym}s) automatically. We define the task, highlight key desired aspects of a good {\acronym}, and propose a model that satisfies these. Finally, we provide an evaluation by human annotators of our generated questions compared against human generated ones, demonstrating competitive performance.

\end{abstract}

\section{Introduction}

Natural language dialogues are often driven by information gaps. Formally, these are gaps between the epistemic states of the interlocutors. Namely, one knows something that the other does not, and the conversation revolves around reducing this gap. An important example is the education setting where teachers ask students questions, and receive answers that may be incomplete. With the expectation of what a {\em complete} answer should contain, the teacher then engages in a gap-focused dialogue to help the student to arrive at a complete answer. There are multiple other application settings of information gaps, including support-line bots, long-form Q\&A, and automated fact checking. 

The core challenge in this setting is how to 
generate effective questions about the information gap.
In terms of  formal semantics and pragmatics, this gap can be viewed as the complementary of the {\em common-ground} \cite{stalnaker2002} held by the interlocutors.  
Somewhat surprisingly, despite much work on dialogue learning \cite{ni2022recent,zhang2020recent} and  question generation   \cite{michael2018qamr,pyatkin-etal-2020-qadiscourse,pyatkin-etal-2021-asking,ko2020inquisitive}, 
little attention has been given to generating questions that focus on such information gaps.

The formal traditional approach to representing the dialogic information gap is via the set of propositions that are known to one side but not the other  \cite{stalnaker2002}. However, this set can be quite large, and it is also unclear how to turn these propositions into dialogue utterances. We propose an arguably more natural representation: a generated set of natural language questions whose answers represent the information that the teacher needs to ask about to reduce the gap. We call these {\em gap-focused questions} ({\acronym}s). A key advantage of this representation is that the generated questions can be used directly in the teacher-student dialogue. 

Given a complete teacher answer and a partial student answer, there are many questions that could be asked, but some are more natural than others. For example, consider the complete answer \textit{``A man is wearing a blue hat and a red shirt and is playing a guitar''}, and a student response \textit{``There is a man playing the guitar''}. 
Two candidate questions could be \textit{``What color hat is the man wearing?''} and \textit{``What is the man wearing?''}. The second question is arguably more natural as it does not reveal information that is not in the teacher-student common ground, namely that a hat is being worn.

The above demonstrates some of the complexity of generating effective {\acronym}s, and the need to rely on certain discourse desiderata.
In this work we define the \acronym challenge,  a novel question generation task,  and we detail the desired properties of the generated questions. Subsequently, we provide a model for {\acronym} generation that aims to satisfy these desiderata, and demonstrate its competitiveness via a task of generating questions to fill the gap between premises and hypotheses in a standard {\em natural language inference} (NLI) setup. 

In designing desired properties for {\acronym}s, we take inspiration from theories of collaborative communication, and in particular Grice's maxims \cite{grice1975logic}. For example, the {\em maxim of  quantity} states that speakers are economic and do not communicate what is already known. Thus, the teacher should not ask about what is already in the common ground with the student. 
In the above example, this means not asking \textit{``What is the man playing?''}. 
We describe additional desiderata in  \S\ref{sec:criteria}.
\commentout{
Furthermore, according to Grice's {\em maxim of quality}, one tries to be truthful, and does not give information for which one lacks adequate evidence. In our setup this requires the teacher to ask only about facts that are known to the them. e.g., do not ask \textit{``which song is the man singing?''}.
}

\begin{figure}[t!]
\vspace{-0.15in}
\hspace{-0.16in}
\includegraphics[width=1.07\columnwidth]{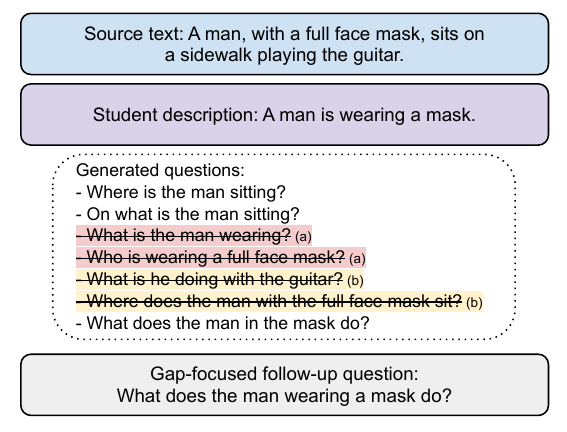} \\
\ \vspace{-0.45in} \\
\caption{Our Gap-Focused Question setup and approach. A student is asked to describe a source text from memory. The goal is to ask a follow-up question about information the student missed. Our approach is to generate a list of candidate questions, and then filter out ones that are either answerable from the student text (red strike-through (a)) or contain facts unknown to the student (yellow strike-through (b)). The follow-up question can be any of the remaining questions.
}
\label{fig:teaser}
\vspace{-0.05in}
\end{figure}

To tackle the GFQ challenge, we show how general-purpose NLP models (question generation, question answering, and constituency parsing) can be used to generate {\acronym}s that satisfy the discourse desiderata. See Figure \ref{fig:teaser} for an outline of the process.
To assess our model, we consider pairs of texts that contain information gaps, and evaluate our ability to capture these gaps using {\acronym}s. Such texts are readily available in NLI datasets that contain pairs of a premise and an entailed hypothesis with less information. We consider the SNLI dataset \cite{bowman2015large}, and use human annotators to evaluate the merit of our approach relative to {\acronym}s generated by humans.

Our contribution is three-fold. First, we propose the novel setup of gap-focused questions, a key element of a student-teacher discourse as well as other settings such as automated fact checking. Second, we identify desiderata inspired by conversational maxims, and provide a model for generating questions that satisfy them. Third, we demonstrate the merit of our model on an NLI dataset.


\section{Related work}
Natural dialogue is a key goal of modern NLP and, despite substantial progress, there is still a considerable difference between humans and models. In this work we focus on dialogues where the bot (teacher) knows more than the user (student), and the goal is to gradually decrease this knowledge gap via gap-focused follow-up questions.

Several works have focused on the problem of follow-up question generation in dialogues. However, to the best of our knowledge, none of these focus on information gaps as we do.
\citet{ko2020inquisitive} introduce the problem of inquisitive question generation, where the goal is to generate questions about facts that are not in the text. This is not done in reference to a complete text, and is thus principally different from our goal. In fact, in our settings, an inquisitive question would typically be a bad {\acronym}, since it refers to information that is outside the knowledge of both teacher and student. Prior works considered a related task referred to as answer-agnostic question generation \cite{scialom2019self}, but with a focus on factual questions, whereas the inquistive setting is broader. 

Another class of follow-up questions are clarification ones \cite{rao2018learning}, which can also be viewed as a special case of inquistive questions. Again, there is no reference to a complete text that defines the information gap. Finally, there are works on follow-up questions guided by rules as in the SHARC dataset \cite{saeidi2018interpretation}. 

Our {\acronym} setting is also related to the challenge of explainable NLI \cite{kalouli-etal-2020-xplainli}, namely the task of explaining why a certain sentence entails another. The {\acronym} output can be viewed as a novel explanation mechanism of why the student text is entailed by the source text, as it explicitly refers to the gap between these texts. 

Our work is inspired by novel uses of question generation models, particularly in the context of evaluating model consistency  \cite{honovich2021q2}. In these, question generation is used to find ``LLM hallucinations'' where the generated text is not grounded in a given reference text. Our task  can be viewed as the inverse of the knowledge grounding task, and our particular focus is on the questions generated rather than just pointing to information gaps. An additional line of work in this vein is QA-based semantics, where text semantics are represented via a set of questions rather than a formal graph \citep[e.g., see][]{michael2018qamr}.

\section{Criteria for Gap-Focused Questions}
\label{sec:criteria}
\begin{figure}[t!]
\vspace{-0.15in}
\hspace{-0.16in}
\includegraphics[width=1.07\columnwidth]{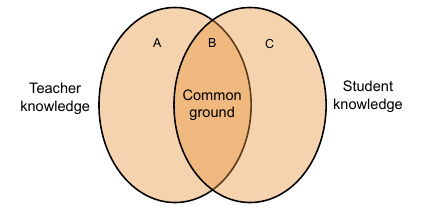} \\
\ \vspace{-0.45in} \\
\caption{In our setup we consider the gaps between the teacher's knowledge (represented by the complete source text $T_c$, areas $A \cup B$ in the diagram) and the student's knowledge (represented by the student text $T_s$, areas $B \cup C$ in the diagram). We consider the information overlap between these two texts as the common ground between the teacher and the student (area $B$), which is a key component in defining good GFQs.
}
\label{fig:venn}
\vspace{-0.05in}
\end{figure}

Given a complete source text $\tc$ and a student text $\ts$, our goal is to construct a model that takes $\ts$ and $\tc$ as input and produces a set of one or more questions $Q$ that ask about the information gap between $\tc$ and $\ts$. If one takes the term ``information gap'' literally, there are many such possible questions (e.g., which word appears in $\tc$ but not in $\ts$). In a natural language setting we are obviously interested in questions that are {\em natural}, that is, would likely be asked by a human who knows $\tc$ and has heard the student description $\ts$. When defining the desiderata for the generated questions, we consider what knowledge is held by the teacher and the student and what information is inside and outside their common ground (see Figure~\ref{fig:venn}). We next identify desired properties for the generated questions, followed by a description of our model for generating gap-focused questions that satisfy these desiderata.

The following desired properties of an effective GFQ are loosely based on collaborative communication concepts \cite{grice1975logic}:
\begin{itemize}[leftmargin=0.2in]
    \item {\bf P1: Answerability:} Only ask questions that can be answered based on the complete text $\tc$ (areas $A\cup B$ in Figure \ref{fig:venn}). {This follows from Grice's {\em maxim of relevance};  speakers say things that are pertinent to the discussion.}

    \item {\bf{P2: Answers should not be in the common ground:}} If the student has already demonstrated knowing a fact in $\ts$, there is no reason to ask about it again. Namely, in Figure \ref{fig:venn}, we don't want to ask about information in $B$. {This pertains to Grice's {\em maxim of quantity}; speakers are economic,  they  do not utter information beyond the bare minimum that is necessary to ask the question, and  they will refrain from repeating already-known information.}

    \item {\bf{ P3: Questions should only use information known to the user:}} The question itself should rely only on information in $\ts$ and not in $\tc$. For example if $\tc$ is \textit{``A Woman is wearing a blue hat''} and $\ts$ is \textit{``A woman is wearing something''}, it is preferable not to ask \textit{``What color is the hat?''} as it refers to information that did not appear in $\ts$ (i.e., that the woman is wearing a hat). This is loosely related to the Grice maxim of manner, where one tries to be clear, brief, and orderly. If we were to ask questions using information unknown to the user (in area $A$ in figure \ref{fig:venn}), we may introduce unnecessary details and obscurity into the discussion.\footnote{Note that in some cases this may only be partially possible and a ``hint'' must be provided in order to be able to phrase a grammatically correct and semantically sensible question.} 

\commentout{
 \item {\bf P4: Do not be too specific.} Asking about information that is too specific has two key shortcomings. First, it may obscure an information-gap that is broader and is of greater importance. Second, it is more likely the student will not be able to answer an overly specific detail. This again follows from Grice's maxim of manner. 
 }
\end{itemize}

\section{The GFQs Generation approach \label{sec:model}}
We next describe our modeling approach for the GFQ generation problem, with the goal of capturing the properties described above.

Before describing our {\acronym}s generation approach, we briefly outline the NLP components we rely on in the question generation process: 
\vspace{0.1in} \newline
{\bf{A question generation model}} $G$ that, given an input text $T$ and a span $X\subset T$, generates questions about $T$ whose answer is $X$.
\commentout{
For example:\\
{\small
{\bf Input}: ``The dog and cat are brown''; ``brown''\\
{\bf Output}: \{``What color is the dog?'',``What color is the cat?'', ``What color are the dog and cat?''\} 
}
}
\vspace{0.05in}\newline
 {\bf{A question answering model}} $A$, that takes as input a text $T$ and a question $Q$ about the text, and returns the answer or an indication that the question is unanswerable from the text.
\commentout{
For example: \\
{\small
{\bf Input}: ``The dog is brown'';``What color is the dog?''\\
{\bf Output}: ``brown'' \\
{\bf Input}: ``The dog is brown''; ``What is the dog's name?'' \\
{\bf Output}: ``UNANSWERABLE''
}
}
\vspace{0.05in}\newline
{\bf{A constituency parser}} $P$, that takes a text $X$, breaks it down into sub-phrases (constituents), and returns a parse tree. \vspace{0.05in} \newline
Additional details about these components can be found in appendix \ref{app:models}.  \vspace{0.1in}
\commentout{
\begin{figure}
    \centering
    \scalebox{0.9}{
    \Tree[.S [.NP DT\\the NN\\dog ] [.VP V\\was [.ADJP ADJ\\brown CC\\and ADJ\\tall ] ] ]}
    \caption{A Parse Tree by a Constituency Parser}
    \label{fig:tree}
\end{figure}
}

We are now ready to describe our approach for generating {\acronym}s. The model generates an ordered set of possible follow-up questions $\genq$ via the following steps, which roughly correspond to the desired criteria described in \S\ref{sec:criteria}:
\vspace{0.1in} \newline
     {\bf Step 1: Generate answerable questions (P1).} Using the constituency parser $P$, we extract the spans of all the constituents in the source text $\tc$, except for those spanning the entire sentence, and single word spans containing functional elements (e.g., prepositions). For each span $X\subset \tc$, we use the question generation model $G$ to generate a set of questions whose answer should be $X$, thus creating a set of questions that satisfy the answerablity property. We denote this set $Q_T$ and assign $\genq=Q_T$.
 \vspace{0.1in}    \newline 
    {\bf Step 2: Filter questions whose answers are in the common ground. (P2).} We next wish to remove questions that are answerable by the student text $\ts$. To that end, we use the question answering model $A$, and for each $q\in\genq$ if $A(\ts, q) \neq \text{``UNANSWERABLE''}$, we set $\genq = \genq \setminus \{q\}$.\footnote{Note that Step 2 will also filter out questions that the student answered incorrectly. This would be an area for improvement in future models.}
\vspace{0.1in} \newline
    {\bf Step 3: Prefer questions which only use information known to the user (P3).} We prefer questions that do not reveal information beyond what is known to the user. This is not always strictly possible and thus, instead of filtering, we rank questions according to the (possibly zero) amount of additional information they reveal. To do so, let $R$ be all the answers to the questions in $Q_G$. By construction $R$ contains spans from $\tc$ that the student didn't mention, i.e. these are spans that we would prefer not to appear in the generated questions. For each $q\in\genq$, we count the number of items in $R$ included in $q$. We sort $\genq$ in ascending order by this number and return the first element. We thus return a question that uses the least number of facts unknown to the student.

\section{Experiments \label{sec:experiments}}
%
We next describe an evaluation of our {\acronym} model.

\paragraph{Data:} We use the {\em SNLI Dataset} \cite{bowman2015large} where a Natural language inference (NLI) pair contains two sentences denoting a premise and a hypothesis, and the relation between them can be \textit{entailment}, \textit{contradiction} and \textit{neutral}. 
We focus on pairs labeled as entailment, and filter out those with bi-directional entailment, so that there is a gap between hypothesis and premise.
\commentout{
We thus filter bi-directional entailments using an NLI model (similar to  \cite{honovich2022true}). In the resulting set of one-directional entailments, the information in the premise ($\tc$) is \emph{strictly} greater than the information in the hypothesis ($\ts$). 
}
We do not use any data for training, and apply our model to the test partition of the SNLI dataset.

\paragraph{Evaluation Benchmark:}
In order to compare the quality of our automatically generated questions to manually generated ones, we asked human annotators to generate questions for 200 instances of the SNLI test set (see Appendix \ref{sec:app_guidelines} for the annotator instructions). We emphasize that these questions were only used for evaluation, as explained below, and not for training the model. They were collected after model design was completed. We release this evaluation dataset to the public, it is available \href{https://storage.googleapis.com/gresearch/gap-focused-questions/data.zip}{here.} See additional details about this dataset in appendix \ref{app:dataset}.

\paragraph{Annotator Evaluation of Generated Questions:} As with other generative settings, offline evaluation is challenging. In fact, even if we had human generated questions for all SNLI, using those for evaluation would need to assume that they are exhaustive (otherwise the model can generate a good question but be penalized because it is not in the set generated by humans). Instead, as is commonly done \cite{ko2020inquisitive}, we rely on human evaluation. We present annotators with $\tc,\ts$ and a candidate \acronym $q$ and ask them to provide a $1-5$ score of how well $q$ functions as a follow-up question (see Appendix \ref{sec:appendix2} for annotators instructions). We use 3 annotators per question.

\begin{table}[t]
\centering
\setlength{\tabcolsep}{5pt} 
{
\begin{tabular}[b]{l c}
    \hline
    Model & Average score \\
    \hline
    Step 1 & 3.72 \\
    Step 2 & 3.86 \\
    Step 3 & 3.94 \\
    Human & 4.06 \\
\end{tabular}
}
\caption{Average scores of the different generation methods on 200 questions, each rated by 3 annotators.}
\label{tab:average_scores} \vspace{-0.05in}
\end{table}

\begin{table*}[]
\resizebox{\textwidth}{!}{%
\begin{tabular}{p{6cm}|p{6cm}|p{6cm}}
\textbf{Source text} & \textbf{Student description} & \textbf{Generated question (Step 3)}\\
\midrule
A man stands by two face structures on Easter Island. & A man on Easter Island.	& Two faces are what on Easter Island? \\
Two young children, one wearing a red striped shirt, are looking in through the window while an adult in a pink shirt watches from behind. & A person in a shirt. & What is one child wearing?\\
A man in a purple jersey is falling down while chasing a player in a green jersey playing soccer & The two soccer players run around chasing each other & What is the man in the cartoon wearing?\\
\end{tabular}%
}
\caption{Examples of the loss patterns found in the analysis of low scoring questions. See details in the Error Analysis paragraph in section \ref{sec:experiments}.}
\label{tab:losses}
\end{table*}

\paragraph{Compared Models:}
We compare four generation approaches:
{\bf Human}: Questions generated by human annotators;
{\bf Step 1}: This model selects a random question out of those generated by the question generation model (i.e., Step 1 in \S\ref{sec:model}). We note that this is already a strong baseline because its questions are based on the source text.
{\bf Step 2}: The outcome of Step 2 in \S\ref{sec:model} where only questions not answerable by the student text are kept. {\bf Step~3}: The outcome of Step 3, where we additionally aim for questions which use information known to the user.

\paragraph{Results:} Table~\ref{tab:average_scores} provides the average scores for each of the considered models and the human generated questions. It can be seen that each step contributes to the score, and human generated questions are somewhat better than our final model ({\bf{Step 3}}). Using the Wilcoxon signed-rank test for paired differences, we found that all differences were significant at p-value $\leq 0.05$.

\commentout{
\paragraph{Experiment 1: Comparison to Baselines.}
Table~\ref{tab:average_scores} provides the average scores for each model. It can be seen that human generated questions are the best, followed by our full model and the random selection baseline. To assess the statistical significant of the differences between the different models, we apply the non-parametric Wilcoxon signed-rank test on the score difference on each pair of generation methods. We find that our approach is superior to the strong baseline with p-value $5.8e-04$, whereas, as expected, the human generation dominates with p-value $6.8e-08$. 
}

\begin{figure}[t!]
\includegraphics[width=1.1\columnwidth]{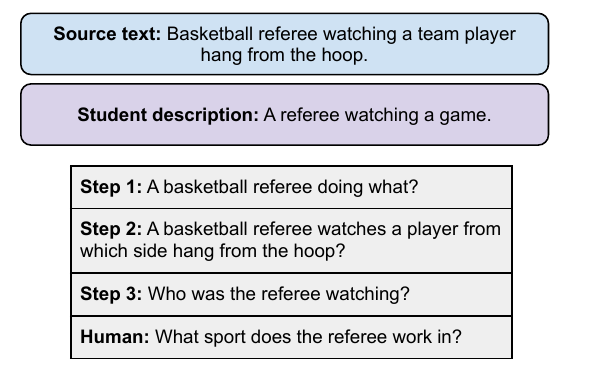} 
\caption{An example of the steps of our Gap-Focused Questions model, and a human-generated question.
}
\label{fig:example}
\vspace{-0.05in}
\end{figure}


\commentout{
\begin{table}[t]
\centering
\footnotesize
\setlength{\tabcolsep}{1pt} 
\begin{tabular}[b]{lcccc}
\hline
	Model & Winner & Left score & Right score & p-value \\
    \hline
    Full vs. Baseline & Full & 3.89 & 3.68 & 5.80e-04 \\
    Human vs. Baseline & Human & 4.18 & 3.68 & 2.06e-16 \\
    Human vs. Full & Human & 4.18 & 3.89 & 6.82e-08 \\
    \hline
\end{tabular}
\caption{Pairwise comparison of our models. The p-value is computed with the Wilcoxon test on the pairs of ratings of each annotator. } 
\label{tab:pairwise_comparison}
\end{table}
}

\paragraph{Examples:} Figure \ref{fig:example} shows an example of the three stages, and a human generated question. Appendix \ref{app:examples} provides more examples.

\paragraph{Error Analysis:} We analyze cases where our final model (Step 3) received low scores from the annotators (an average score of 3 and lower). In our analysis we have observed three main loss patterns (sometimes appearing together): {\bf{(1)}} Poor question phrasing --- these are questions whose structure or choice of words is less natural than if a person were to ask the same question. See example in the first row in Table \ref{tab:losses}. {\bf{(2)}} Questions which include information outside of the teacher-student common ground. These are cases where the minimum criterion defined in Step 3 still results in a question with some information unknown to the user. See examples in the first 2 rows in Table \ref{tab:losses}. {\bf{(3)}} Questions including information outside the complete source text. In rare cases we have found that the question generation model generates questions that include ``hallucinations'' or point to issues in the semantic understanding of the complete source text. See the third example in Table \ref{tab:losses}.

\section{Conclusion}
We consider the task of question generation in a novel setting where there is an information gap between speakers, and the gap-focused questions ({\acronym}s) aim to reduce this gap. Building on advances in question generation and question answering, we show how to generate useful {\acronym}s that meet several natural criteria inspired by theories cooperative  conversation. 

It is natural to ask whether one can employ a fully generative approach for GFQs using LLMs. This is a natural direction for future study, and we believe that the criteria and design choices we studied here will be significant in defining and evaluating  such future work.

\section*{Limitations}
We present the first study of generating questions for filling in information gaps. Our method is limited in several ways. First, it focuses on information that is explicitly missing, and does not discuss information that is inaccurate or incomplete in other ways. Second, it only asks one follow-up question and does not address multi-turn dialogue about a student answer, or multiple student answers. Finally, our approach makes somewhat restricted use of the student answer, and it will be better to generate questions that directly uptake information from the student text \cite{demszky2021measuring}. We leave the deep investigation of these for future work.

\section*{Acknowledgments}
We thank Avi Caciularu for constructive feedback on this work.

\section*{Ethics and Impact}
Regarding risks, as with any NLP model, care must be taken in application, so that it generates truthful information, and does not introduce biases. However, we think this is not a major concern in our case as our modeling will generate text directly related to the source and student texts. In terms of impact,  our approach can be used to improve a wide array of applications, including educational dialogue (e.g., reading comprehension), support-line bots, and automated fact checking. 
\bibliography{main}
\bibliographystyle{acl_natbib}

\clearpage
\appendix

\section{Annotating Guidelines}
\label{sec:app_guidelines}
Here we provide all the guidelines to annotators, for both human question generation 
and human rating of questions generated by the model.

\begin{figure*}
    \centering
    \includegraphics[scale=.5]{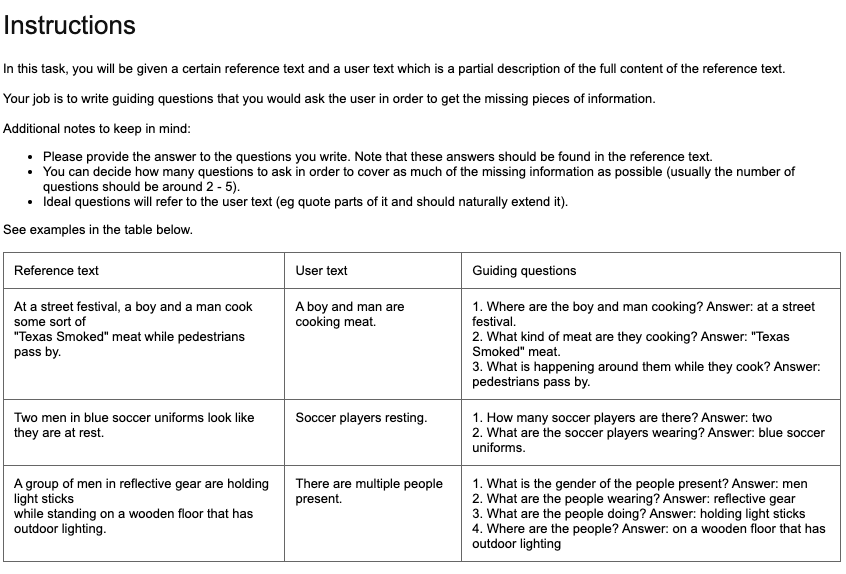}
    \caption{Human annotator guidelines and examples for the task of writing follow-up questions.}
    \label{fig:my_label_1}
\end{figure*}

\begin{figure*}
    \centering
    \includegraphics[scale=.45]{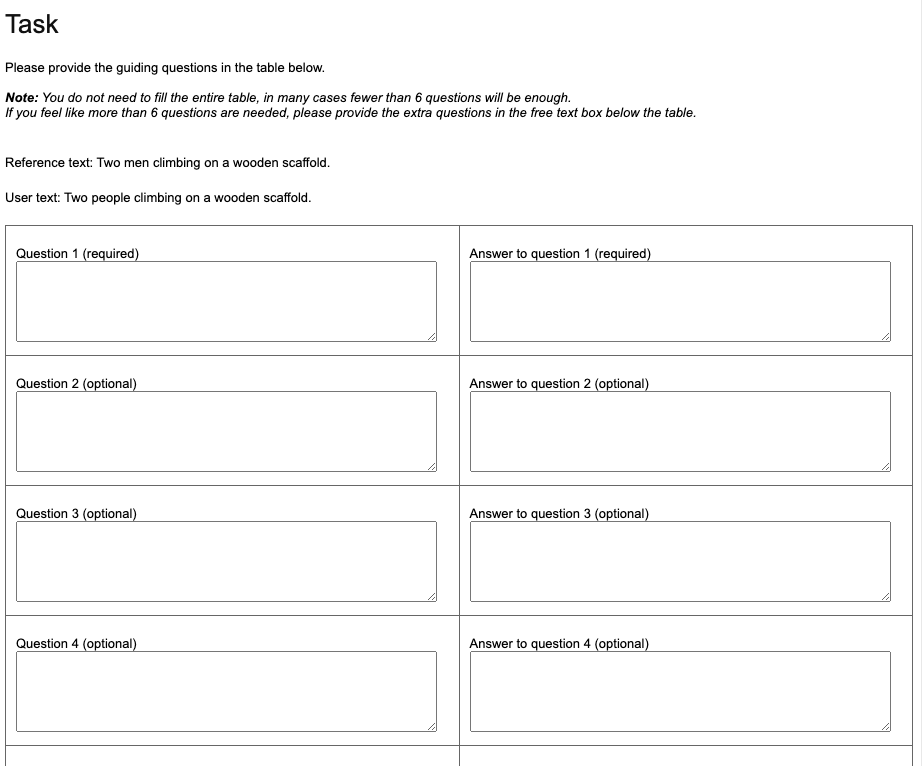}
    \caption{The user interface of the human annotator task of writing follow-up questions.}
    \label{fig:my_label_2}
\end{figure*}

\paragraph{Guidelines for the human annotator task of writing follow-up questions:}
We depict the guidelines and the examples for the writing follow-up questions task in Figure~\ref{fig:my_label_1}, and the task design in Figure~\ref{fig:my_label_2}.

\begin{figure*}
    \centering
    \includegraphics[scale=.3]{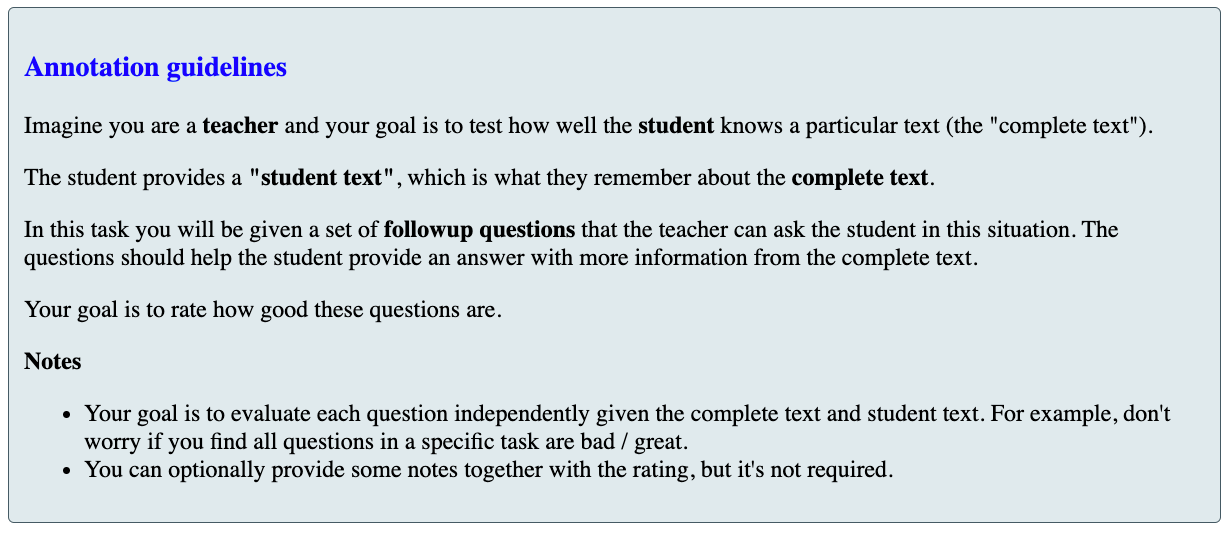}
    \caption{Guidelines for the human annotator task of rating follow-up questions.}
    \label{fig:my_label_3}
\end{figure*}

\begin{figure*}
    \centering
    \includegraphics[scale=.3]{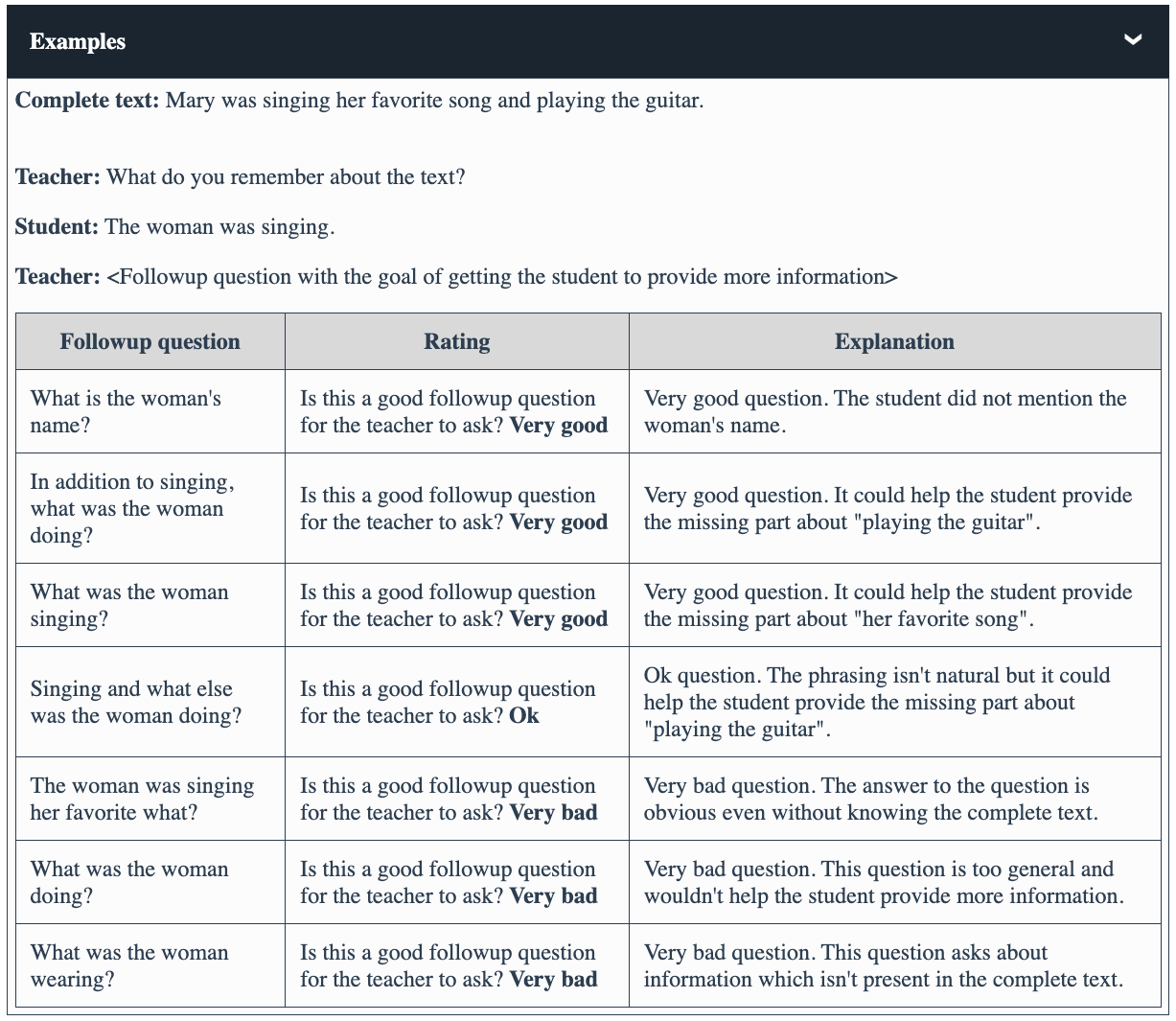}
    \caption{Task examples (that are originally attached to the guidelines) for the task of rating follow-up questions.}
    \label{fig:my_label_4}
\end{figure*}

\begin{figure*}
    \centering
    \includegraphics[scale=.22]{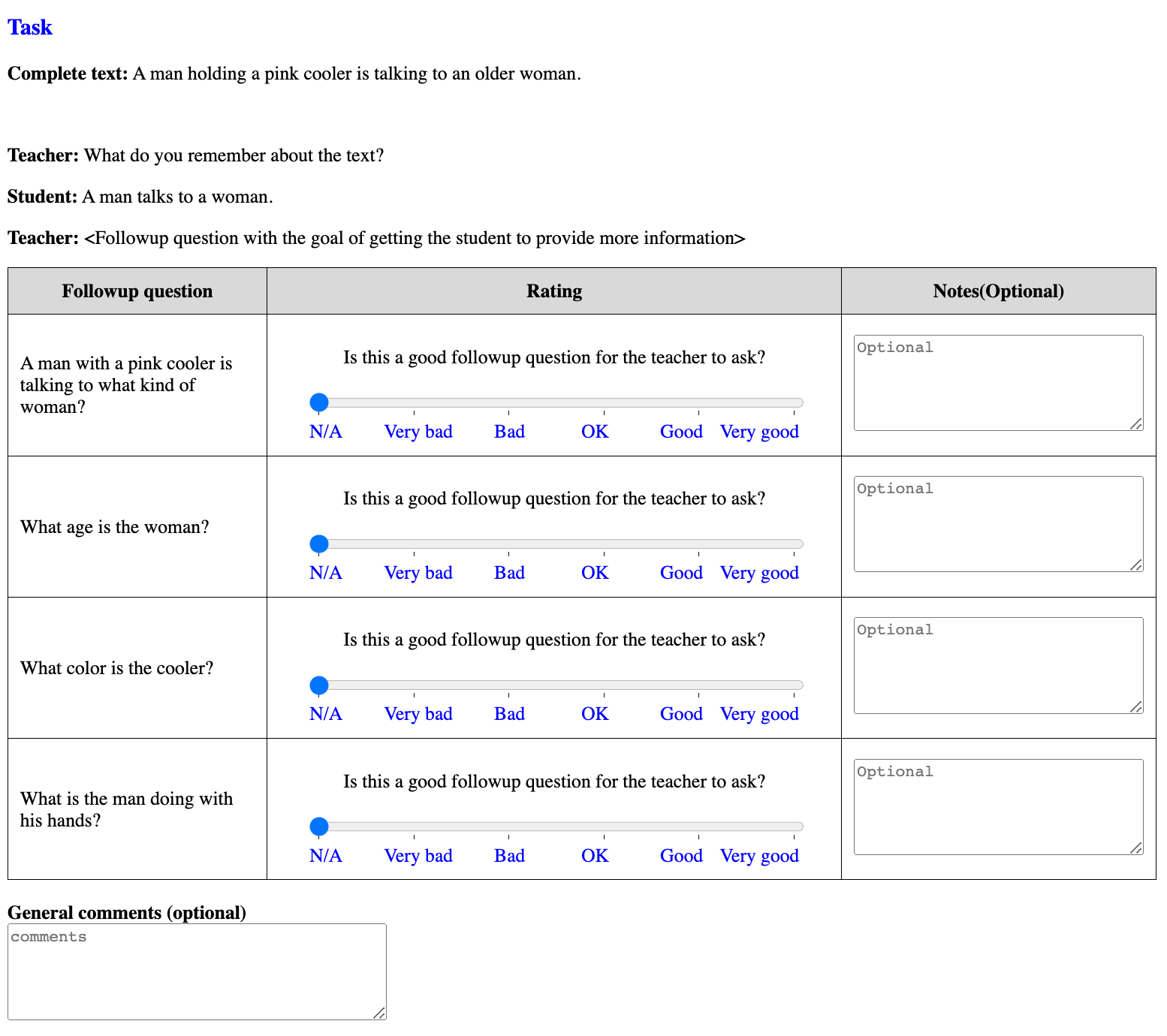}
    \caption{The user interface of the human annotator task of rating follow-up question.}
    \label{fig:my_label_5}
\end{figure*}

\paragraph{Guidelines for the human annotator task of rating follow-up questions:}
We depict the guidelines of the task of rating the follow-up questions in Figure~\ref{fig:my_label_3}, the examples in Figure~\ref{fig:my_label_4}, and the task design in Figure~\ref{fig:my_label_5}.

\label{sec:appendix2}

\section{Annotator Related Information}
Annotators were paid by the hour, and recruited as contractors for a variety of annotating projects by our team and related teams. The annotators are all native English speakers (from Canada and the US). They are also aware of the way in 
which the information will be used. There are no special ethical sensitivities in the collection process and thus it was exempt from an ethics review board.

\section{Implementation Details}
\label{app:models}
\paragraph{Question Generation Model:} As our question generation model $G$, we use the T5-xxl model \cite{T5} fine-tuned on SQuAD1.1 
\cite{rajpurkar-etal-2016-squad}. We also use beam search and question filtering, similarly to \citet[Section 2]{honovich2021q2}, see this work for further details.
\paragraph{Question Answering Model:} For our question answering model $A$, we use the T5-xxl model \cite{T5} fine-tuned on SQuAD2.0 \cite{rajpurkar-etal-2018-know}.
\paragraph{Constituency Parser:} We use the Berkeley Neural Parser~\cite{kitaev-klein-2018-constituency}, implemented in the spaCy package.\footnote{We used spaCy3.0 -- \url{https://spacy.io/}.}
\paragraph{SNLI Filtering:} We consider the subset of SNLI with an ``entailed'' label. 
Since we are not interested in the case of equivalent hypothesis and premise, we filter out  bi-directional entailments using an NLI model (similar to  \cite{honovich2022true}). In the resulting set of one-directional entailments, the information in the premise ($\tc$) is \emph{strictly} greater than the information in the hypothesis ($\ts$), which is our case of interest. 

\section{Computational Resources Details}
In terms of computational resources, the project is lightweight, as it required no training at all, and just running inference steps of pre-trained models (question answering, question generation and parsing), all of which run in several minutes on standard GPUs. 

\section{GFQ test released dataset}
\label{app:dataset}
We release a benchmarking dataset of 200 examples from SNLI test with a human generated gap-focused question. The data is available \href{https://storage.googleapis.com/gresearch/gap-focused-questions/data.zip}{here.}

\paragraph{Details about the dataset}
We asked 3 annotators to write questions for each SNLI pair (see guidelines in appendix \ref{sec:app_guidelines}) and used a heuristic to select a single GFQ. When selecting this single question our goal is to prefer GFQs where multiple annotators chose to write a question about the same topic. We therefore apply the following heuristic: for each human written question $q$ we used our question answering model $A$ and define $a$ as the answer to this question given $T_c$: $a = A(T_c, q)$. We then count $n$: the number of annotators which produced questions leading to the same answer $a$, we look at the questions for which $n$ is maximal and choose a random question from there.

\paragraph{License} This data as well as the underlying SNLI data are licensed under a Creative Commons Attribution-ShareAlike 4.0 International License \footnote{http://creativecommons.org/licenses/by-sa/4.0/}.
 
\section{Examples of Generated Questions}
\label{app:examples}
Here we provide examples of questions generated by humans and by the different models we consider.
Table \ref{tab:examples_exp1} reports questions generated by Step 1, Step 2, Step 3 and Human.

\begin{table*}[t!]
\centering
\resizebox{\textwidth}{!}{%
\begin{tabular}{p{6cm}|p{12cm}}
\toprule {}
    \textbf{Source text} & \textbf{A child plays with her father's boots.}\\
    \textbf{Student description} & \textbf{A child is playing.} \\
    \midrule
    Step 1 & What does she do with them? \\
    Step 2 & What does the child do with her father's boots? \\
    Step 3 & What does the child play with? \\
    Human & What is the child playing with? \\
    \midrule\midrule
    \textbf{Source text} & \textbf{Two men work outside polishing shoes.} \\
    \textbf{Student description} & \textbf{Some men are polishing shoes.} \\
    \midrule
    Step 1 & What are the two men doing to the shoes? \\
    Step 2 & Who works outside to polish shoes? \\
    Step 3 & Where do the men work? \\
    Human & How many men are there? \\
    \midrule\midrule
    \textbf{Source text} & \textbf{A boy dressed in a plaid kilt with a brown hat wields a long pole.} \\
    \textbf{Student description} & \textbf{A boy has and object in his hands.} \\
    \midrule
    Step 1 & Aside from the kilt, what brown item does the boy wearing it wear? \\
    Step 2 & What color is the hat the boy is wearing? \\
    Step 3 & What type of garment is the boy wearing? \\
    Human & What does the boy wear on his body? \\
    \midrule\midrule
    \textbf{Source text} & \textbf{A man in a white shirt and baseball hat is pushing a cart carrying several bags on a street.} \\
    \textbf{Student description} & \textbf{A man is walking outside.} \\
    \midrule
    Step 1 & What is the man pushing a cart wearing?  \\
    Step 2 & Where is the man pushing a cart with bags?	 \\
    Step 3 & What is the man in the picture wearing? \\
    Human & What is the man wearing? \\
    \bottomrule
\end{tabular}}
\caption{Example GFQs from our different models: Step 1, Step 2, Step 3 and Human.}
\label{tab:examples_exp1}
\end{table*}

\section{Data Related Information}
The data collected from annotators contains the manually generated questions and the scoring of generated questions. There are no issues of offensive content or privacy in this data, as it based closely on the SNLI dataset.
\end{document}